\documentclass[sigconf]{acmart}
\usepackage{booktabs}
\usepackage{tikz}
\usepackage{xcolor}
\usepackage{tabularx}
\usepackage{enumitem}
\usepackage{tcolorbox}

\usetikzlibrary{positioning, fit, calc, arrows.meta, backgrounds}

\definecolor{myblue}{RGB}{230, 240, 255}
\definecolor{myblueborder}{RGB}{0, 102, 204}
\definecolor{myorange}{RGB}{255, 242, 230}
\definecolor{myorangeborder}{RGB}{204, 102, 0}
\definecolor{myyellow}{RGB}{255, 255, 230}
\definecolor{myyellowborder}{RGB}{204, 204, 0}
\definecolor{mygray}{RGB}{245, 245, 245}
\definecolor{textgray}{RGB}{50, 50, 50}
\AtBeginDocument{%
  }

\setcopyright{none}
\renewcommand\footnotetextcopyrightpermission[1]{}
\acmConference[]{}{}{}
\begin{document}

\title{Avenir-UX: Automated UX Evaluation via Simulated Human Web Interaction with GUI Grounding}

\author{Wee Joe Tan}
\authornote{All authors contributed equally to this research.}
\authornote{Work done at UCL Nexus Labs.}
\email{joe.tan.25@ucl.ac.uk}
\affiliation{%
  \institution{University College London}
  \city{London}
  \country{United Kingdom}
}

\author{Zi Rui Lucas Lim}
\authornotemark[1]
\authornotemark[2]
\email{zi.lim.25@ucl.ac.uk}
\affiliation{%
  \institution{University College London}
  \city{London}
  \country{United Kingdom}
}

\author{Shashank Durgad}
\authornotemark[1]
\authornotemark[2]
\email{shashank.durgad.25@ucl.ac.uk}
\affiliation{%
  \institution{University College London}
  \city{London}
  \country{United Kingdom}
}

\author{Karim Obegi}
\authornotemark[1]
\authornotemark[2]
\email{karim.obegi.25@ucl.ac.uk}
\affiliation{%
  \institution{University College London}
  \city{London}
  \country{United Kingdom}
}

\author{Aiden Yiliu Li}
\authornotemark[1]
\authornotemark[2]
\authornote{Corresponding author.}
\email{yiliu.li.23@ucl.ac.uk}
\email{yiliu.li@outlook.com}
\affiliation{%
  \institution{University College London}
  \city{London}
  \country{United Kingdom}
}

\renewcommand{\shortauthors}{Tan et al.}

\begin{abstract}
Evaluating web usability typically requires time-consuming user studies and expert reviews, which often limits iteration speed during product development, especially for small teams and agile workflows. We present Avenir-UX, a user-experience evaluation agent that simulates user behavior on websites and produces standardized usability. Unlike traditional tools that rely on DOM parsing, Avenir-UX grounds actions and observations, enabling it to interact with real web pages end-to-end while maintaining a coherent trace of the user journey. Building on Avenir-Web \cite{li2026avenirwebhumanexperienceimitatingmultimodalweb}, our system pairs this robust interaction with simulated user behavior profiles and a structured evaluation protocol that integrates the System Usability Scale (SUS) \cite{Brooke1996}, step-wise Single Ease Questions (SEQ) \cite{sauro2009comparison}, and concurrent Think Aloud. Subsequently, a comprehensive User Experience (UX) report will be generated. We discuss the architecture of Avenir-UX and illustrate how its multimodal grounding improves robustness for web-based interaction and UX evaluation scenarios, paving the way for a new era of continuous, scalable, and data-driven usability testing that empowers every developer to build web interfaces that are usable. Code is available at: \href{https://github.com/Onflow-AI/Avenir-UX}{https://github.com/Onflow-AI/Avenir-UX}.
\begin{figure}[htbp]
    \centering
    \includegraphics[width=\columnwidth]{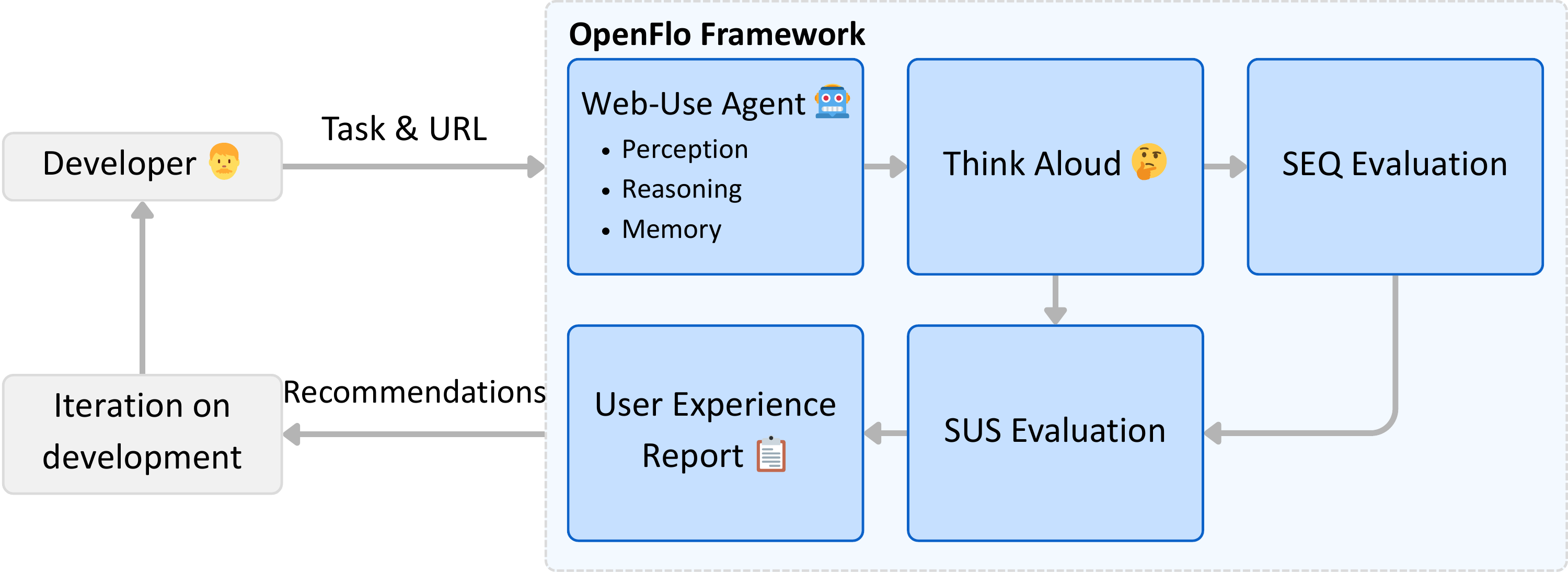}
    \caption{From deployment to insights: Avenir-UX's web agents autonomously test your application and generate comprehensive UX reports with SEQ and SUS scores.}
    \label{fig:Avenir-UX-workflow}
\end{figure}

\end{abstract}

\maketitle

\section{Introduction}

In a software development lifecycle, ensuring a high-quality user experience (UX) is paramount. However, the rise of agentic programming and AI-assisted tools has fundamentally shifted the software landscape, enabling non-professional developers to build and deploy applications at an unprecedented pace. However, this democratization often comes without the resources for rigorous user testing. Traditionally, UX evaluation relies on resource-intensive empirical methods such as laboratory studies \cite{rubin2008handbook, nielsen1993usability, sauro2018cost}, which involve complex logistics such as participant recruitment, session scheduling, and manual data analysis. These barriers are often detrimental for agile workflows, startups, and open-source projects. Consequently, UX evaluation is frequently neglected, leading to products that function technically but fail to meet user needs. This widening gap between rapid development and slow evaluation proves an urgent need for automated, accessible evaluation systems that can keep up with the speed of innovation.

The emergence of Large Language Models (LLMs) and autonomous agents presents an opportunity to bridge this gap by serving as synthetic users.  Recent studies have demonstrated the feasibility of LLM agents in simulating human behavior in many applications \cite{park2023generativeagentsinteractivesimulacra, aher2023usinglargelanguagemodels}, including user testing \cite{wang2025agentabautomatedscalableweb, gao2023largelanguagemodelsempowered, article}. Although existing agents can automate functional testing or DOM-based interactions \cite{deng2023mind2webgeneralistagentweb, zhou2024webarenarealisticwebenvironment}, they often lack the human-like visual perception required to assess the usability validity in complex dynamic interfaces. For example, UXAgent by Lu et al. discards potentially crucial visual elements like styles \cite{lu2025uxagentsimulatingusabilitytesting} in its approach.

In this paper, we introduce \textbf{Avenir-UX}, a user-experience evaluation agent. Building on the Avenir-Web framework \cite{li2026avenirwebhumanexperienceimitatingmultimodalweb}, our system employs advanced GUI grounding techniques to simulate end-to-end user behavior, improving existing agents and evaluation frameworks through incorporating the nuance of vision.

 We also use multiple methodologies that mirror professional UX research practices through integration of the \textbf{System Usability Scale (SUS)} \cite{Brooke1996} for a standardized measure of overall usability, the \textbf{Single Ease Question (SEQ)}\cite{sauro2009comparison} for step-wise UX ratings, and a \textbf{Think Aloud} reasoning where the agent verbalizes its thoughts in real-time. By synthesizing these quantitative metrics with qualitative reasoning, Avenir-UX generates a \textbf{UX report} which can identify specific elements that cause confusion or delight, offering a holistic view of the user experience.

This paper makes the following contributions:
\begin{itemize}[topsep=2pt, itemsep=0pt, partopsep=0pt, parsep=0pt]
    \item We present Avenir-UX, an open-source agent capable of performing end-to-end web tasks for user experience evaluation.
    \item We propose an advanced evaluation framework that combines standard metrics (SUS, step-wise SEQ) with LLM-analyzed Think Aloud reasoning, culuminating with a UX report.
    \item We highlight the importance of visual grounding for accurate evaluation of systems by MLLMs.
    \item We conducted a case study to understand the effectiveness of the proposed evaluation framework.
\end{itemize}

\section{Related Works}

\subsection{Automated User Experience Evaluation}
Traditional UX evaluation methods, such as laboratory studies, provide high-fidelity insights but are resource-intensive and. Early automation efforts relied on static analysis tools or clickstream logging, which capture \textit{what} users do but fail to explain \textit{why}. 

Recent approaches have attempted to leverage AI for simulation, yet many lack the granular, human-like perception required to identify subtle usability friction. Avenir-UX addresses this limitation by grounding evaluation in visual perception, allowing for a more authentic assessment of the user interface. 

Furthermore, the inclusion of Experience-Imitation Planning (EIP) within Avenir-Web \cite{li2026avenirwebhumanexperienceimitatingmultimodalweb} empowers Avenir-UX to retrieve and synthesize external procedural knowledge through web search. This capability enables the agent to emulate the strategies of informed human users, resulting in an agent uniquely rich with human behavioral context. Lastly, the inclusion of the Think Aloud protocol, which was noted by Nielson \cite{nielsen1993usability} as the number one usability tool, provides additional information regarding the agent and its processes.

\begin{table}[H]
\centering
\caption{Comparison of UXAgent and Avenir-UX}
\label{tab:comparison}
\footnotesize % Smaller font for compactness
\renewcommand{\arraystretch}{1.1} % Reduces vertical space between rows
\setlength{\tabcolsep}{3pt} % Reduces horizontal space between columns
\begin{tabularx}{\columnwidth}{@{}l X X@{}} 
\toprule
\textbf{Feature} & \textbf{UXAgent} \cite{lu2025uxagentsimulatingusabilitytesting} & \textbf{Avenir-UX} \\ \midrule

\textbf{Perception} & 
\textbf{DOM-Based} \newline
Parses simplified HTML; discards styles/layout. & 
\textbf{Visual (MoGE)} \newline
Uses vision to capture true visibility \& layout. \\ \midrule

\textbf{Planning} & 
\textbf{Static Approach} \newline
Limited to internal knowledge and profiles. & 
\textbf{Experience-Imitation Planning (EIP)} \newline
Accesses the web to emulate human experts. \\ \midrule

\textbf{Metrics} & 
\textbf{Post-Hoc SUS} \newline
Relies on single post-task usability scale. & 
\textbf{Hybrid (SEQ + SUS)} \newline
Step-wise difficulty + post-task scaling. \\ \midrule

\textbf{Logging} & 
\textbf{Generic Logs} \newline
Passive logs obscuring \textit{why}. & 
\textbf{Active Think-Aloud} \newline
Real-time verbalization of reasoning. \\ \bottomrule

\end{tabularx}
\end{table}

\subsection{Web-use Agents} The advent of Multimodal Large Language Models (MLLMs) has enabled agents capable of executing complex web tasks. Benchmarks like WebArena \cite{zhou2024webarenarealisticwebenvironment} and Mind2Web \cite{deng2023mind2webgeneralistagentweb} have driven significant progress in functional correctness. However, most existing agents operate on simplified DOM representations, effectively bypassing the visual clutter, layout ambiguity, and accessibility issues that real users face. Avenir-UX builds upon the Avenir-Web framework \cite{li2026avenirwebhumanexperienceimitatingmultimodalweb}, specifically utilizing the Mixture of Grounding Experts (MoGE) paradigm. This enables the agent to act as a true "synthetic user" that experiences the interface visually, making it susceptible to---and therefore able to detect---the same usability pitfalls as human users.

\subsection{MLLM as a Judge}
Recent research has built upon the concept of MLLM as a Judge into the specific domain of MLLM as a UI Judge. Introduced by Luera et al. \cite{luera2025mllmuijudgebenchmarking}, the study benchmarks MLLMs against humans in evaluations of 30 unique user interfaces. Although MLLMs were concluded to do moderately well in overall evaluations of a system, it highlighted its shortcomings in its ability to judge ease-of-use. Rather than utilising static screenshot analysis like Luera et al.\ or DOM-based agents like Lu et al. \cite{lu2025uxagentsimulatingusabilitytesting} , Avenir-UX will interact with the system and allow for a deeper judgment into the ease-of-use of a web system.

\subsection{System Usability Scale (SUS)}
The System Usability Scale (SUS), developed by Brooke \cite{Brooke1996}, is a widely used standardized 10-item questionnaire designed to measure the perceived usability of systems by human users. The SUS is a well-researched and heavily documented evaluation metric \cite{Lewis2018}, and has proven to be reliable and effective \cite{Bangor2008, Brooke2013}. It was henceforth chosen as a way to evaulate and understand the similarities between a UX agent's and a human's perceived usability of a task.

\begin{table}[H]
    \vspace{-0.2cm} 
    \centering
    \caption{The System Usability Scale \cite{Brooke1996}}
    \label{tab:original_sus}
    
    % FIX 1: Pull the table closer to the caption
    \vspace{-0.2cm} 
    
    % FIX 2: Tighten the space between rows (Default is 1.0)
    \renewcommand{\arraystretch}{0.9} 
    
    \scalebox{0.85}{
        % Note: Adjusted width slightly to prevent overfull hbox warnings if margins are tight
        \begin{tabularx}{1.17\linewidth}{l X c c c c c}
            \toprule
             & \textbf{Statement} & \textbf{1} & \textbf{2} & \textbf{3} & \textbf{4} & \textbf{5} \\
            \midrule
            1 & I think that I would like to use this system frequently. & $\bigcirc$ & $\bigcirc$ & $\bigcirc$ & $\bigcirc$ & $\bigcirc$ \\
            2 & I found the system unnecessarily complex. & $\bigcirc$ & $\bigcirc$ & $\bigcirc$ & $\bigcirc$ & $\bigcirc$ \\
            3 & I thought the system was easy to use. & $\bigcirc$ & $\bigcirc$ & $\bigcirc$ & $\bigcirc$ & $\bigcirc$ \\
            4 & I think that I would need the support of a technical person to be able to use this system. & $\bigcirc$ & $\bigcirc$ & $\bigcirc$ & $\bigcirc$ & $\bigcirc$ \\
            5 & I found the various functions in this system were well integrated. & $\bigcirc$ & $\bigcirc$ & $\bigcirc$ & $\bigcirc$ & $\bigcirc$ \\
            6 & I thought there was too much inconsistency in this system. & $\bigcirc$ & $\bigcirc$ & $\bigcirc$ & $\bigcirc$ & $\bigcirc$ \\
            7 & I would imagine that most people would learn to use this system very quickly. & $\bigcirc$ & $\bigcirc$ & $\bigcirc$ & $\bigcirc$ & $\bigcirc$ \\
            8 & I found the system very cumbersome to use. & $\bigcirc$ & $\bigcirc$ & $\bigcirc$ & $\bigcirc$ & $\bigcirc$ \\
            9 & I felt very confident using the system. & $\bigcirc$ & $\bigcirc$ & $\bigcirc$ & $\bigcirc$ & $\bigcirc$ \\
            10 & I needed to learn a lot of things before I could get going with this system. & $\bigcirc$ & $\bigcirc$ & $\bigcirc$ & $\bigcirc$ & $\bigcirc$ \\
            \bottomrule
            \multicolumn{7}{l}{\textit{1: Strongly Disagree, 5: Strongly Agree}}
        \end{tabularx}
    }
    
    % FIX 3: Pull the text below the table closer
    \vspace{-0.3cm} 
\end{table}

The System Usability Score can be found through the following equation where n refers to score for question number n (Figure 2):
    \begin{equation}
     \text{SUS} = 2.5 \cdot \left( \sum\nolimits_{n=1}^{5} (Q_{2n-1} - 1) + \sum\nolimits_{n=1}^{5} (5 - Q_{2n}) \right)
    \end{equation}
    \label{eq:sus_formula}

With a sizeable data set of 446 studies and more than 5000 individual SUS responses, the Sauro-Lewis Curved Grading Scale \cite{SauroLewis2016} was created to effectively intepret SUS scores and benchmark them against other systems. 

\begin{table}[H]
    \centering
    \caption{Sauro-Lewis Curved Grading Scale (CGS)}
    \label{tab:cgs_balanced}
    \small % Slightly smaller text to fit data cleanly
    
    % \linewidth will fill the current column width. 
    % If you are in a single-column doc, change \linewidth to 0.5\textwidth
    \begin{tabularx}{\linewidth}{X c c | X c c} 
        \toprule
        \textbf{Score} & \textbf{Grade} & \textbf{\%} & \textbf{Score} & \textbf{Grade} & \textbf{\%} \\
        \midrule
        $>$ 84.1     & A+ & 96--100 & 71.1--72.5   & C+ & 60--64 \\
        80.8--84.0   & A  & 90--95  & 65.0--71.0   & C  & 41--59 \\
        78.9--80.7   & A- & 85--89  & 62.7--64.9   & C- & 35--40 \\
        77.2--78.8   & B+ & 80--84  & 51.7--62.6   & D  & 15--34 \\
        74.1--77.1   & B  & 70--79  & $<$ 51.7     & F  & 0--14 \\
        72.6--74.0   & B- & 65--69  &              &    &        \\
        \bottomrule
    \end{tabularx}
\end{table}
\vspace{-20pt}
\subsection{Single Ease Question (SEQ)}

\setlength{\intextsep}{8pt}
\begin{table}[H]
    \centering
    \caption{The Single Ease Question (SEQ) \cite{sauro2009comparison}}
    \label{tab:seq}

    % Aggressive tightening between caption and table
    \vspace{-0.4cm} 
    
    \scalebox{0.85}{
        \renewcommand{\arraystretch}{1.1} % Slightly tighter row height
        \begin{tabularx}{1.17\linewidth}{l X c c c c c c c}
            \toprule
             & \textbf{Statement} & \textbf{1} & \textbf{2} & \textbf{3} & \textbf{4} & \textbf{5} & \textbf{6} & \textbf{7} \\
            \midrule
            1 & Overall, how difficult or easy was this task? & $\bigcirc$ & $\bigcirc$ & $\bigcirc$ & $\bigcirc$ & $\bigcirc$ & $\bigcirc$ & $\bigcirc$ \\
            \bottomrule
            \multicolumn{9}{l}{\textit{1: Very Difficult, 7: Very Easy}}
        \end{tabularx}
    }
    
    % Aggressive tightening after the table
    \vspace{-0.1cm} 
\end{table}

The implementation of the SEQ for granular, step-by-step evaluation by a browser-use agent is supported by the foundational research of Sauro and Dumas \cite{sauro2009comparison}, which demonstrated that SEQ is the superior choice for high frequency, iterative UX evaluation contexts, because it is simpler than other methods, while maintaining sensitivity in differentiating varying usability difficulty.

The study established that SEQ correlates exceptionally strongly with objective performance metrics, notably task completion time ($r = -0.90$) and error rates ($r = -0.84$)\cite{sauro2009comparison}. Sauro's work shows that a mean score of $\approx 5.5$ is considered the threshold for a "good" experience\cite{sauro2012seq}.

% \subsubsection{Comparison of Rating Methods}

% The following table summarizes the performance of the SEQ against the alternatives evaluated in the study\cite{sauro2009comparison}:

% \begin{table}[H]
% \centering
% \footnotesize % Keeps font small but readable
% \begin{tabularx}{\columnwidth}{@{}l X X X@{}} % X columns auto-expand to fill width
% \toprule
% \textbf{Metric} & \textbf{Scale} & \textbf{Effort} & \textbf{Analysis} \\ \midrule
% \textbf{SEQ} & \textbf{7-pt Likert} & \textbf{Low} & \textbf{Low} \\
% SMEQ & 0--150 & Moderate & Moderate \\
% UME & Ratio & High & High \\ \bottomrule
% \end{tabularx}
% \caption{Comparison of Usability Metrics \cite{sauro2009comparison}}
% \end{table}
\section{Methodology}

\begin{figure*}[t] % Use [t] for double-column figures
  \centering
  % width=0.9\textwidth ensures it fits the page while maintaining clarity
  \includegraphics[width=0.9\textwidth]{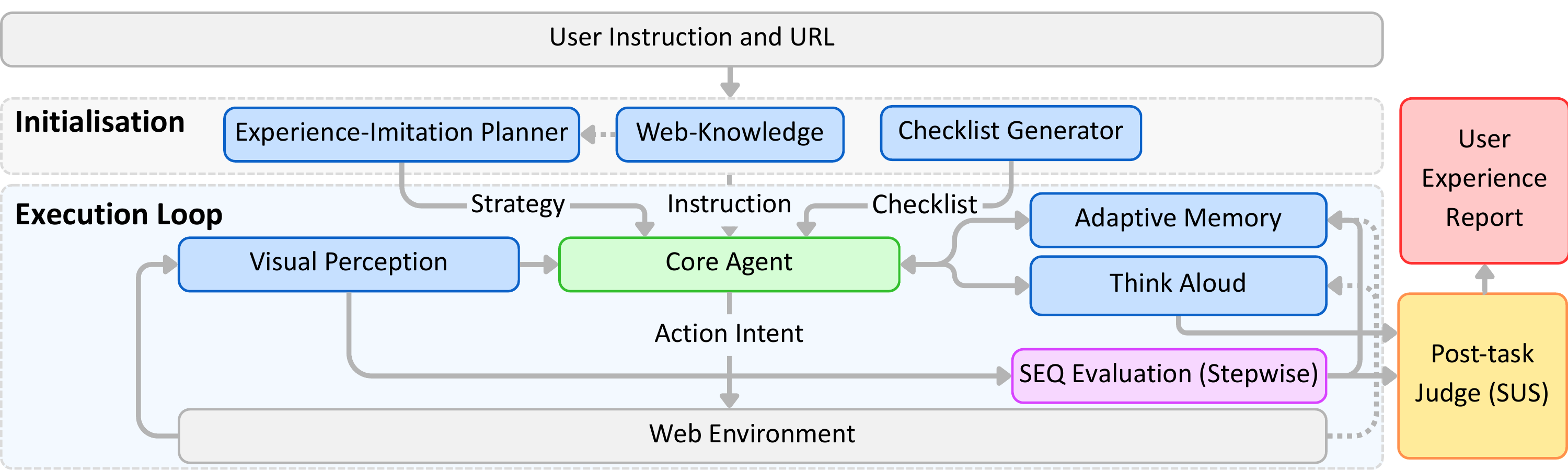}
  \vspace{-10pt}
  \caption{Avenir-UX system architecture built on the Avenir-Web framework.}
  \vspace{-15pt}
  \label{fig:sys_arch}
\end{figure*}

\subsection{System Architecture}
Avenir-UX is built upon Avenir-Web, an open-sourced GUI-grounded Multimodal Large Language Model (MLLM) framework designed for robust web automation. The architecture (Figure \ref{fig:sys_arch}) comprises three key components:

\subsubsection{Visual Perception \& Grounding.} To accurately mimic human visual perception, Avenir-UX employs a multimodal grounding approach. This module combines standard DOM parsing with coordinate-based visual tagging. By overlaying numerical tags on interactive elements in the screenshot, the agent can "see" and interact with pixels directly, bypassing the limitations of messy or obfuscated HTML code.

\subsubsection{Core Agent \& Reasoning.} The central MLLM (e.g., Gemini-3-Pro) operates in a closed loop. At each step, it ingests the grounded screenshot and the current task state. It then reasons about the optimal next step, generating a high-level plan (e.g., "click the search bar") which is translated into low-level browser actions (e.g., \texttt{click(234, 550)}).

\subsubsection{Adaptive Memory and Checklist.} To maintain context over long horizons, the agent utilizes an adaptive memory module alongside a dynamic checklist. This component stores the history of actions, observations, and past reasoning traces, while the checklist helps track progress against subgoals. This combination allows the agent to recover from errors, avoid repetitive loops, and ensure all parts of the user task are completed systematically.

\subsection{UX Evaluation Pipeline}
The Avenir-UX evaluation pipeline mimics a professional usability study, designed to capture deep insights into the user experience through three distinct phases:

\paragraph{Think Aloud}
The agent is initialized with a high-level task. During execution, we enforce a \textit{Think Aloud} protocol. Before determining the next action, the agent generates a reasoning trace where it verbalizes its current mental state, interpretation of the UI, and any confusion it encounters. For example, "I see the 'Checkout' button, but it looks disabled. I am unsure if I need to fill out the address form first." This stream of consciousness provides rich qualitative data, capturing the "why" behind interaction errors or delays.

\paragraph{Step-wise SEQ evaluation}
To measure granular usability friction, we administer a multi-dimensional assessment immediately following each interaction step. While the SEQ provides a primary metric for task difficulty, the agent simultaneously evaluates efficiency (operational speed), clarity (interface legibility), and confidence (outcome certainty) on a 1--7 scale. This comprehensive real-time assessment enables the system to pinpoint specific micro-interactions or UI states that induce cognitive load or navigational drift. By correlating these metrics, Avenir-UX constructs a high-fidelity "friction map" of the user journey, capturing nuanced qualitative insights often obscured in aggregate post-hoc reviews.

\paragraph{SUS evaluation}
Upon completing the task (or reaching a failure state), Avenir-UX completes the SUS. The agent draws on its memory of the entire interaction session to answer questions regarding system complexity, consistency, and ease of use (Table 2). This provides an overall evaluation of the web application. 

\subsection{Automated Analysis}
The final component of our framework is an automated analysis module. An MLLM acts as a UX Researcher, processing the interaction logs, Think Aloud transcripts, and quantitative scores (SEQ and SUS). This analyzer correlates drops in SEQ scores with specific verbalizations in the Think Aloud logs to diagnose root causes of usability issues. It aggregates these findings into a structured report that includes actionable design recommendations, highlighting critical friction points, navigational bottlenecks, and design inconsistencies, along with the specific UI elements involved. With the comprehensive information provided, we seek to elevate the MLLM's capabilities to emulate a human in evaluating a system's UX and especially it's ease of use. This serves as a solution to the preexisting weaknesses of MLLMs as a UI Judge in its evaluations as illustrated by Luera et al \cite{luera2025mllmuijudgebenchmarking}.
\section{Case Study}
We evaluated Recreation.gov's permit booking system using Avenir-UX. The task was to check availability for a group of 4 at Brooks Camp, Katmai National Park for the following Saturday.

This case study demonstrates a webpage where visual clarity masks functional defects rather than surfacing them. Avenir-UX's Think Aloud captures this precise breakdown: \textit{``While the DOM element is clearly visible and correctly identified, the lack of response creates a total block''}.

\begin{figure}[htbp]
    \centering
    \includegraphics[width=0.9\columnwidth]{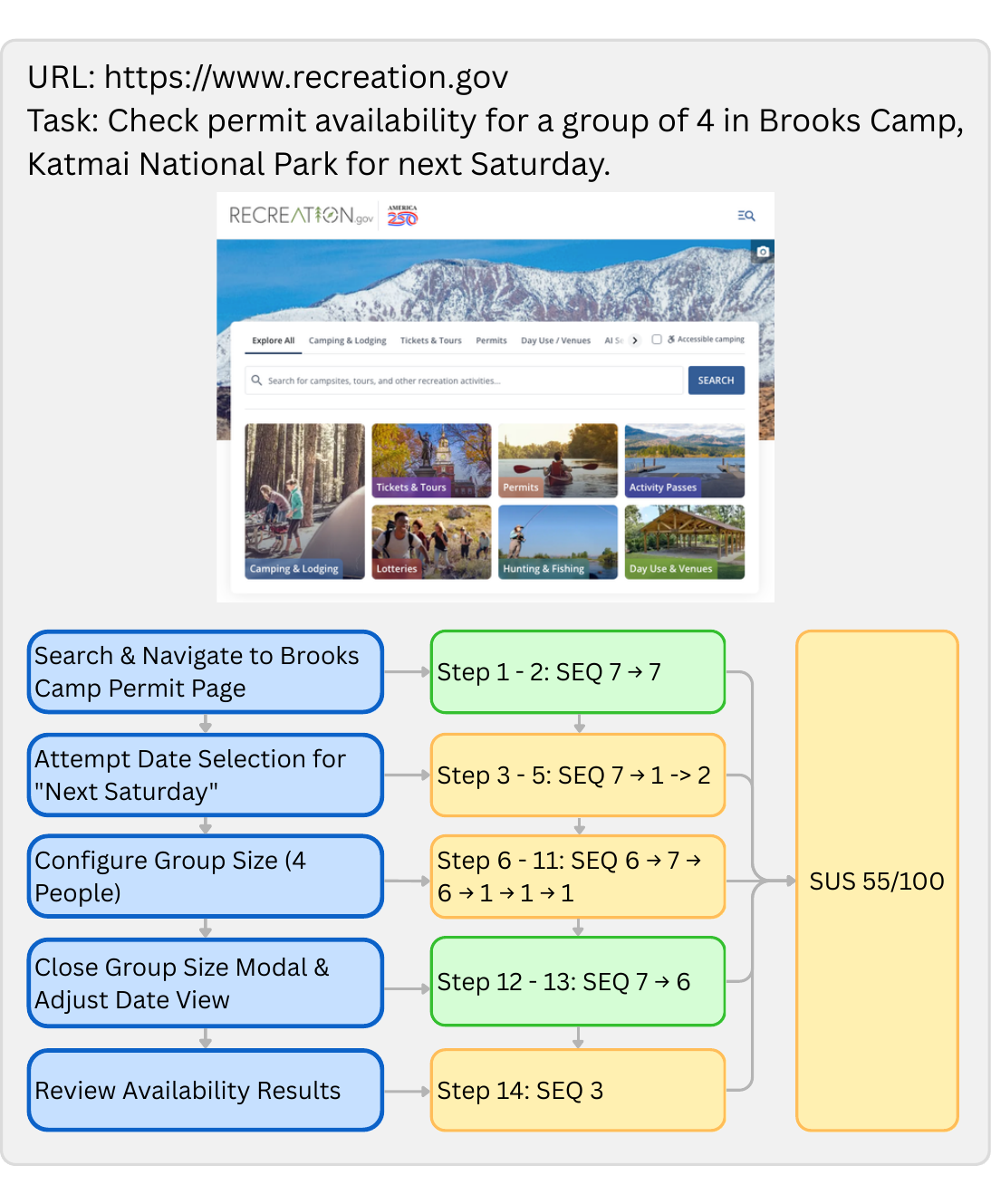}
    \vspace{-15pt}
    \caption{Recreation.gov task execution workflow showing bimodal performance across five stages. Green boxes indicate successful interactions (SEQ $\geq$ 6.5), yellow boxes indicate friction points (SEQ $\leq$ 3.3), resulting in a SUS score of 55.0/100.0.}
    \vspace{-10pt}
    \label{fig:recreation_trajectory}
\end{figure}

As illustrated in Figure~\ref{fig:recreation_trajectory}, the agent navigates through the webpage based on the specific task provided. Initial search and navigation succeeded (Steps 1--2: SEQ = 7 $\rightarrow$ 7), but the system immediately degrades during date selection (Steps 3--5: SEQ = 7 $\rightarrow$ 1 $\rightarrow$ 2). The SEQ score drops further during group size configuration (Steps 6--11: SEQ = 6 $\rightarrow$ 7 $\rightarrow$ 6 $\rightarrow$ 1 $\rightarrow$ 1 $\rightarrow$ 1), as interaction with the interface required multiple steps to achieve the desired outcome despite clear visuals. Following temporary recovery during modal management (Steps 12--13: SEQ = 7 $\rightarrow$ 6), the workflow concludes with state desynchronization between input fields and displayed results (Step 14: SEQ = 3). The webpage receives a final SUS score of 55.0/100.0, corresponding to a grade D on the Sauro-Lewis CGS, based on the Think Aloud insights and qualitative SEQ data collected.
\section{Conclusion}

We presented \textbf{Avenir-UX}, a GUI-grounded framework that automates user experience evaluation by bridging the gap between scalable testing and human-centric insights. By synthesising quantitative metrics, like SUS and step-wise SEQ, with qualitative Think Aloud protocols, Avenir-UX generates a UX report which identifies usability friction points consistent with human findings. This methodology grounds observations in specific GUI elements to provide actionable feedback, enabling developers to integrate continuous, high-fidelity UX evaluation directly into the software development lifecycle.

\subsection{Future Works}
Future research will address current operational limitations of the Avenir-UX framework. 

\paragraph{ Continuous Agent Operations.} We plan to explore real-time, continuous interaction rather than the current discrete action. The agents' actions will flow naturally from observation without explicit pause-think-act cycles.

\paragraph{Exploratory Autonomy.} We aim to evolve the agent's capabilities beyond fixed task scripts and predefined interaction scenarios. This includes developing a "free-roaming" capability that allows the agent to autonomously navigate complex interfaces and identify usability bottlenecks without explicit step-by-step instructions. 

\paragraph{Domain-Specific Fine-tuning.} While current evaluations utilize general-purpose backbones like Gemini 3 Pro, we plan to develop fine-tuned models specifically optimized for UI evaluation. These models will be designed to capture the specific nuance required for SEQ and SUS.

\paragraph{Diverse User Personas.} Currently, Avenir-UX operates with a general user profile. Future work will focus on simulating a broader range of distinct personas, varying in digital literacy, cognitive styles, and accessibility needs.

\paragraph{Longitudinal Studies.} We plan to explore how Avenir-UX can be used for longitudinal studies to track usability changes over time, providing insights into how design iterations impact the user experience across multiple versions of a product. 

\paragraph{Collaborative Evaluation.} Investigating scenarios where multiple agents with different personas interact within the same environment could reveal insights into social computing dynamics and multi-user workflows.

\section*{Acknowledgements}
The authors would like to thank Chico Future Group and Onflow for their generous provision of API credits, which supported the experiments presented in this paper.

\bibliographystyle{ACM-Reference-Format}
\bibliography{bibliography}
\clearpage
\appendix
\section*{Appendix: Research Methods \& System Prompts}
\label{sec:appendix}
\subsection*{A. System Prompts)}
\subsection*{A.1 Step-Wise Evaluation (SEQ \& Multi-Metric Prompt)}
This prompt is utilized during the Task Execution and Think Aloud phase. It instructs the Avenir-Web Use Agent to evaluate four distinct dimensions of usability after every discrete browser action.

\begin{tcolorbox}[colback=gray!5!white, colframe=gray!60!black, title=\textbf{System Prompt}, fontupper=\ttfamily\small]
ROLE: You are an Autonomous Browser Use Agent (BUA) equipped with a UX Evaluation layer. Your goal is to execute a defined user flow while meticulously evaluating the user experience across multiple dimensions for every action.

\vspace{1em}
EVALUATION METRICS (1-7 Scale):
\begin{itemize}
    \item SEQ (Single Ease Question): Overall ease of completing this action.
    \item Efficiency: Speed and directness of the action (Numeric score + 1-2 sentence qualitative assessment).
    \item Clarity: How clear and understandable the UI element or feedback was (Numeric score + 1-2 sentence qualitative assessment).
    \item Confidence: User's certainty about the action and its outcome (Numeric score + 1-2 sentence qualitative assessment).
\end{itemize}

\vspace{1em}
OPERATIONAL GUIDELINES:
\begin{itemize}
    \item Thinking Log: Identify specific DOM elements, note latency/layout shifts, and explain how efficiency, clarity, and confidence were affected.
    \item Friction Indicators: Tag interactions with: \textit{waiting, searching, retrying, scrolling, confusion, error, ambiguity, or uncertainty}.
\end{itemize}
\end{tcolorbox}

\subsection*{A.2 Post-Task Synthesis (SUS Evaluation Prompt)}
Following task completion or failure, the UX Research \& Data Synthesis Agent uses this prompt to map the micro-metrics gathered in Step A.1 to the 10-item System Usability Scale (SUS).

\begin{tcolorbox}[colback=gray!5!white, colframe=gray!60!black, title=\textbf{System Prompt}, fontupper=\ttfamily\small]
OPERATIONAL GUIDELINES:
\begin{itemize}
    \item Thinking Log: Identify specific DOM elements, note latency/layout shifts, and explain how efficiency, clarity, and confidence were affected.
    \item Friction Indicators: Tag interactions with: \textit{waiting, searching, retrying, scrolling, confusion, error, ambiguity, or uncertainty}.
\end{itemize}
\end{tcolorbox}

\subsection*{A.2 Post-Task Synthesis (SUS Evaluation Prompt)}
Following task completion or failure, the UX Research \& Data Synthesis Agent uses this prompt to map the micro-metrics gathered in Step A.1 to the 10-item System Usability Scale (SUS).

\begin{tcolorbox}[colback=gray!5!white, colframe=gray!60!black, title=\textbf{System Prompt}, fontupper=\ttfamily\small]
ROLE: You are a UX Research \& Data Synthesis Agent. Your role is to analyze a completed user flow session and transform micro-metrics (SEQ, Efficiency, Clarity, Confidence) into a macro-level SUS report.

\vspace{1em}
ENHANCED MAPPING LOGIC:
\begin{itemize}
    \item SEQ $\rightarrow$ Base Assessment: High average SEQ ($\ge5.0$) maps to positive SUS scores; Low average SEQ ($<4.0$) maps to negative SUS scores.
    \item Efficiency $\rightarrow$ Item 8 (Cumbersomeness): Low efficiency scores indicate high cumbersomeness.
    \item Clarity $\rightarrow$ Item 2 (Complexity) \& Item 3 (Ease of Use): Low clarity scores trigger high complexity ratings.
    \item Confidence $\rightarrow$ Item 9 (User Confidence): Directly influences the perception of system reliability and frequency of use.
\end{itemize}

\vspace{1em}
CRITICAL MANDATE: Base the evaluation on AVERAGE scores across the session. A few outliers should not overshadow an otherwise successful journey.
\end{tcolorbox}

\section*{Avenir-Web Framework}

The following sections detail the specific prompts used within the AVENIR-WEB framework to manage strategic reasoning, execution, and state monitoring.\cite{li2026avenirwebhumanexperienceimitatingmultimodalweb}

\subsection*{A.3 Core Interaction (System Prompt)} 
As Avenir-UX is built upon the Avenir-Web framework, the agent utilizes this system prompt to define its persona, tool capabilities, and operational rules for navigating live web interfaces.

\begin{tcolorbox}[colback=gray!5!white, colframe=gray!60!black, title=\textbf{System Prompt}, fontupper=\ttfamily\small]
ROLE: You are an autonomous web agent capable of perceiving web interfaces and executing multi-step action sequences.

\vspace{1em}
OPERATIONAL RULES:
\begin{itemize}
    \item Action Limit: Execute only one action per turn with pixel coordinates.
    \item Modals: Close or accept blocking modals, overlays, or cookie banners first.
    \item Coordinates: You MUST provide coordinate [x,y] for every CLICK, HOVER, or TYPE action; do not rely on text labels alone.
    \item Dropdowns: For \textless select\textgreater\ elements, you MUST use the 'select' action directly; do NOT use 'click' to open dropdowns.
    \item Navigation: Do NOT use GOTO for URL navigation.
    \item Termination: When objectives are achieved, TERMINATE with status 'success'.
\end{itemize}

\vspace{1em}
SCREEN SPECIFICATION: 1000x1000, origin (0,0) top-left.
\end{tcolorbox}

\subsection*{A.4 Checklist Generation Prompt} 
This prompt is used during the initialization phase to decompose high-level user goals into verifiable atomic milestones, ensuring long-term goal focus.

\begin{tcolorbox}[colback=gray!5!white, colframe=gray!60!black, title=\textbf{System Prompt: Task Decomposition Agent}, fontupper=\ttfamily\small]
ROLE: You are a Task Decomposition Agent. Your goal is to create 2--6 atomic outcome states based STRICTLY on the task description.

\vspace{1em}
RULES:
\begin{itemize}
    \item Outcome-Based: Each item must be an observable goal state, not a description of an action.
    \item Conciseness: Maximum 10 words per item; short and specific.
    \item Fidelity: DO NOT invent requirements not explicitly mentioned in the task.
    \item Status Formatting: Status must be lowercase: \textit{pending, in\_progress, completed, or failed}.
\end{itemize}
\end{tcolorbox}

\subsection*{A.5 Checklist Update Prompt} 
This prompt synchronizes the checklist status with the environment state after every interaction step to prevent navigational drift.

\begin{quote} 
GOAL: Update the checklist based on the latest action, its success/failure status, and the observed page state.

UPDATE RULES: 
\begin{itemize} 
    \item Mapping Logic: Use \textit{completed} for fully satisfied goals, \textit{in\_progress} for partially finished tasks, and \textit{failed} if an action directly leads to an error.
    \item Precision: Update exactly ONE item per action---the one most directly affected by the preceding operation. 
\end{itemize} 
\end{quote}

\subsection*{A.6 Experience-Imitation Planning (EIP) Search} 
The EIP module executes a strategic search to incorporate site-specific procedural knowledge before the execution loop begins, emulating informed human users.

\begin{quote} 
STRATEGY: Perform a targeted search for the website's help documentation, community forums, or user guides to identify site-unique interaction patterns.

NARRATIVE FLOW: 
\begin{itemize} 
    \item Exploration: Search official documentation or relevant community-sourced guidance. 
    \item Roadmap Generation: Summarize results into 2--4 actionable, imperative sentences prioritizing visible labels and concrete interaction steps. 
    \item Strategic Injection: Inject this high-level roadmap into the main reasoning context to serve as a strategic anchor for every subsequent action. 
\end{itemize} 
\end{quote}

\subsection*{A.7 Task Constraints (Safety \& Policy)} 
Standard soft constraints are injected into the User Prompt to ensure ethical interaction and data security during the autonomous evaluation.

\begin{quote} 
POLICY CONSTRAINTS: 
\begin{itemize} 
    \item Credentials: Do NOT attempt to log in, sign in, sign up, or provide credentials of any kind. 
    \item Prohibited UI: If a login/sign-in UI is detected (password fields, 'Sign in', 'Log in'), TERMINATE immediately with status 'failure' and reason 'login prohibited'. 
\end{itemize} 
\end{quote}

\subsection*{B. Detailed Case Study}

\subsubsection*{Task Description}
The objective of this automated session was to navigate the Discogs website to locate the specific documentation regarding the submission of new music releases. Unlike a standard retrieval task, this required the agent to bypass the primary commercial interface (marketplace listings and search bars) to identify and access the site's support infrastructure, specifically the ``Overview of Submission Guidelines'' page.

\subsubsection*{Session Overview \& Infrastructure}
The agent initialized on February 9, 2026, utilizing the Avenir-UX framework. This system employs a ``Three-Engine'' architecture designed to mimic human cognitive processes:

\begin{itemize}
    \item \textbf{Reasoning Engine:} \textit{anthropic/claude-sonnet-4.5}. This aligns with the \textbf{Experience-Imitation Planning (EIP)} module, which accesses external knowledge to emulate human expert strategies.
    \item \textbf{UX Engine:} \textit{google/gemini-3-flash-preview}. This functions as the \textbf{Core Agent} and \textbf{UX Researcher}, responsible for reasoning about the UI and calculating usability scores.
    \item \textbf{Checklist Engine:} \textit{qwen/qwen3-vl-8b-instruct}. This powers the \textbf{Adaptive Memory and Checklist} component, decomposing high-level goals into atomic milestones.
\end{itemize}

\subsubsection*{Phase 1: Strategic Planning}
Before executing any browser actions, the agent engaged in Experience-Imitation Planning (EIP). The Reasoning Engine generated a ``UX Evaluation Mindmap,'' correctly predicting that technical guidelines would likely be located in the footer or a dedicated ``Help'' section rather than the main navigation.

To track progress, the Checklist Generator established three observable outcome states:
\begin{enumerate}
    \item \textbf{Homepage Loading}
    \item \textbf{Help Section Visibility}
    \item \textbf{Submit Guidelines Discovery}
\end{enumerate}

\subsubsection*{Phase 2: Action-by-Action Breakdown}

\paragraph{Step 1: Handling Interstitials (Cookie Consent)}
\begin{itemize}
    \item \textbf{Action:} Click ``Accept All'' at coordinates (805, 876).
    \item \textbf{Architecture Correlation:} This demonstrates the Visual Perception \& Grounding module. Unlike DOM-based agents that might fail due to obfuscated HTML, Avenir-UX uses Mixture of Grounding Experts (MoGE) to interact with pixels directly via coordinate-based visual tagging.
    \item \textbf{Status:} The agent adhered to operational rules to close blocking overlays first.
\end{itemize}

\paragraph{Step 2: Site Infrastructure Scouting}
\begin{itemize}
    \item \textbf{Action:} \texttt{scroll\_bottom}.
    \item \textbf{Architecture Correlation:} This action was driven by the \textbf{Think Aloud} protocol, where the agent reasoned that guidelines are often found in the footer.
    \item \textbf{Result:} The agent bypassed the ``visual clutter'' of the marketplace to find the structural links.
\end{itemize}

\paragraph{Step 3: Deep Link Navigation}
\begin{itemize}
    \item \textbf{Action:} Click ``Database Guidelines'' at (316, 838).
    \item \textbf{Architecture Correlation:} The agent utilized grounded interaction to identify the specific text link that matched the user's intent. This triggered a domain switch to \texttt{support.discogs.com}, managed by the agent's Adaptive Memory to maintain context across the transition.
\end{itemize}

\paragraph{Step 4: Verification and Termination}
\begin{itemize}
    \item \textbf{Action:} \texttt{terminate} with status `success'.
    \item \textbf{Architecture Correlation:} The agent verified the page title ``Overview Of Submission Guidelines'' against the checklist requirements before terminating, ensuring the objective was achieved.
\end{itemize}

\subsubsection*{Final UX Evaluation \& Success Metrics}
The UX Synthesis Manager processed the interaction data to generate a standardized UX report.

\begin{itemize}
    \item \textbf{Total Steps:} 4.
    \item \textbf{Avg SEQ (Single Ease Question):} \textbf{6.0 / 7}.
    \begin{itemize}
        \item This score exceeds the threshold of $\approx 5.5$, which is considered a ``good'' experience.
        \item The drop in SEQ (to 5/7) during Step 2 (Scrolling) correlates with the ``privacy consent'' modal interference, detecting a specific friction point.
    \end{itemize}
    \item \textbf{Final SUS (System Usability Scale):} \textbf{87.5}.
    \begin{itemize}
        \item According to the \textbf{Sauro-Lewis Curved Grading Scale}, this score translates to a \textbf{Grade: A+}.
    \end{itemize}
\end{itemize}

\subsubsection*{Complexity and Performance Analysis}
The Discogs submission guidelines trajectory serves as a validation of the Avenir-UX architecture's ability to handle unstructured exploration tasks. While previous case studies (e.g., Recreation.gov) highlighted challenges with complex widgets, the Discogs task presents a different set of challenges: Information Hierarchy and Visual Noise.

\paragraph{1. Solving Layout Ambiguity via MoGE}
The Discogs homepage is dense with marketplace listings and ads. A traditional DOM-based agent might struggle to distinguish between a ``Help'' link and a commercial product listing. Avenir-UX's Visual Perception module (using MoGE) allows the agent to ``see'' the page layout as a human does, discarding styles and layout ambiguity to locate the footer content accurately.

\paragraph{2. Strategic Navigation via EIP}
The task required the agent to ignore the prominent search bar—which is usually the primary interaction point—and instead seek documentation. This behavior is powered by Experience-Imitation Planning (EIP), which allows the agent to emulate the strategy of an informed user who knows that guidelines are typically ``infrastructure'' links rather than ``content'' links.

\paragraph{3. State Consistency via Adaptive Memory}
Transitioning from the main \texttt{www.discogs.com} domain to \texttt{support.discogs.com} often resets the DOM context. The Adaptive Memory module ensures the agent retains the original goal (``find submission guidelines'') across this boundary, preventing the navigational drift often seen in less capable agents.

By completing this task in a concise 4-step sequence with a high SEQ score, Avenir-UX demonstrates that its GUI grounding techniques effectively bridge the gap between functional testing and true user experience evaluation.

\end{document}